\documentclass{article} 
\usepackage{iclr2026_conference,times}

\iclrfinaltrue

\usepackage{amsmath,amsfonts,bm}

\def\eqref#1{equation~\ref{#1}}

\def\1{\bm{1}}

\DeclareMathAlphabet{\mathsfit}{\encodingdefault}{\sfdefault}{m}{sl}
\SetMathAlphabet{\mathsfit}{bold}{\encodingdefault}{\sfdefault}{bx}{n}

\usepackage{hyperref}
\usepackage{url}
\usepackage{soul}
\usepackage{xcolor}
\usepackage{graphicx}
\usepackage{booktabs}

\usepackage{algorithm}
\usepackage{algpseudocode}
\usepackage{url}            
\usepackage{amsfonts}       
\usepackage{amsmath}
\usepackage{nicefrac}       
\usepackage{microtype}      
\usepackage{amssymb}
\usepackage{breqn}
\usepackage{bbm}
\usepackage{enumerate}
\usepackage{listings}
\usepackage{colortbl}
\usepackage{float} 
\usepackage[most]{tcolorbox}
\usepackage{soul} 
\usepackage{xspace}
\usepackage{makecell}
\usepackage{multirow}
\usepackage{subcaption}
\definecolor{ours}{RGB}{220, 87, 114}
\definecolor{baseline}{RGB}{79, 94, 187}
\usepackage{siunitx}
\usepackage{wrapfig}
\usepackage{pgfplots}
\usepackage{arydshln}
\usepackage{fontawesome5}

\usepgfplotslibrary{groupplots,colormaps}
\pgfplotsset{compat=1.18}

\definecolor{salmon}{RGB}{250, 128, 114}
\definecolor{wheat}{RGB}{245,222,179} 
\definecolor{grass}{RGB}{0,128,0} 
\definecolor{darkpink}{RGB}{255, 20, 147}
\definecolor{darkpurple}{RGB}{75, 0, 130}
\newcommand{\MASK}{\texttt{<MASK>} }
\newcommand{\EOS}{\texttt{<EOS>} }
\newcommand{\nonEOS}{\texttt{<non-EOS>} }
\newcommand{\PAD}{\texttt{<PAD>} }
\newcommand{\Trap}{\sethlcolor{brown!55}\hl{\texttt{<EOS>} Trap!} }

\title{Taming Masked Diffusion Language Models via  Consistency Trajectory Reinforcement Learning with Fewer Decoding Step}

\author{
\qquad \qquad \quad Jingyi Yang\textsuperscript{\rm 1, \rm 2} \quad Guanxu Chen\textsuperscript{\rm 2, \rm 3} \quad Xuhao Hu\textsuperscript{\rm 1, \rm 2} \quad Jing Shao\textsuperscript{\rm 2}\thanks{Corresponding Author.} \\
\qquad \qquad \quad \textsuperscript{\rm 1}Fudan University \quad \textsuperscript{\rm 2}Shanghai Artificial Intelligence Laboratory \\
\qquad \qquad \quad \textsuperscript{\rm 3}Shanghai Jiao Tong University
\\
\qquad \qquad \quad \texttt{yangjingyi946@gmail.com}, \quad \texttt{shaojing@pjlab.org.cn}\\
\qquad \qquad \qquad \qquad \quad \faGithub \, \textbf{Code:} \href{https://github.com/yjyddq/EOSER-ASS-RL}{\textcolor{darkpink}{https://github.com/yjyddq/EOSER-ASS-RL}} \\
}

\begin{document}

\maketitle

\begin{abstract}
Masked diffusion language models (MDLMs) have recently emerged as a promising alternative to autoregressive (AR) language models, offering properties such as parallel decoding, flexible generation orders, and the potential for fewer inference steps. Despite these advantages, decoding strategies and reinforcement learning (RL) algorithms tailored for MDLMs remain underexplored. A naive approach is to directly transfer techniques well-established for AR models to MDLMs. However, this raises an immediate question: Is such a naive transfer truly optimal? For example, 1) Block-wise and semi-AR decoding strategies are not employed during the training of MDLMs—so why do they outperform full diffusion-style decoding during inference? 2) Applying RL algorithms designed for AR models directly to MDLMs exhibits a training-inference inconsistency, since MDLM decoding are non-causal (parallel). This results in inconsistencies between the rollout trajectory and the optimization trajectory.
To address these challenges, we propose \textbf{EOS} \textbf{E}arly \textbf{R}ejection (\textbf{EOSER}) and \textbf{A}scending \textbf{S}tep-\textbf{S}ize (\textbf{ASS}) decoding scheduler, which unlock the potential of MDLMs to perform full diffusion-style decoding, achieving competitive performance with fewer decoding steps. Additionally, we introduce \textbf{C}onsistency Tra\textbf{j}ectory \textbf{G}roup \textbf{R}elative \textbf{P}olicy \textbf{O}ptimization (\textbf{CJ-GRPO}) for taming MDLMs, which emphasizes the consistency between rollout trajectory and optimization trajectory, and reduces the optimization errors caused by skip-step optimization.
We conduct extensive experiments on reasoning tasks, such as mathematical and planning benchmarks, using LLaDA-8B-Instruct. The results demonstrate that the proposed EOSER and ASS mechanisms, together with CJ-GRPO, hold significant promise for effectively and efficiently taming MDLMs. 
\end{abstract}

\section{Introduction}
\label{sec:introduction}

\begin{figure}[t!]
  \centering
    \includegraphics[width=1.\textwidth]{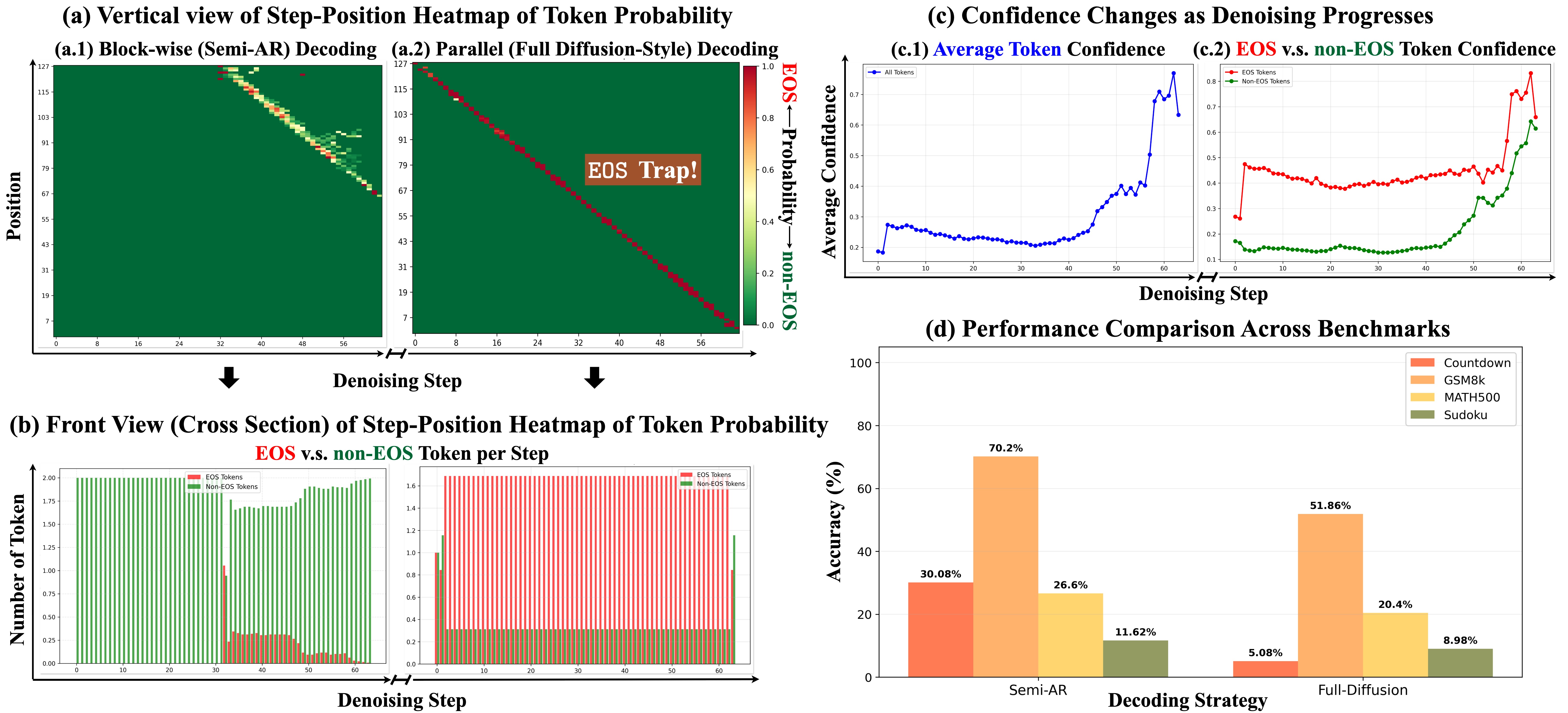}
    \vspace{-0.7cm}
    \caption{\textbf{(a)} and \textbf{(b)} illustrate the vertical and front view of Step-Position Heatmap of Token Probability, showing the frequency of \EOS appearing at each position per step. \textbf{(c)} depicts the performance comparison of semi-AR and full-diffusion decoding strategies. \textbf{(d)} shows the average confidence of tokens as denoising progresses. }
    \label{fig:motivation}
\end{figure}

Autoregressive (AR) large language models~\cite{brown2020language,bai2023qwen,touvron2023llama,grattafiori2024llama,guo2025deepseek,yang2025qwen3} have demonstrated compelling scaling laws in general capabilities, alignment, and reasoning. 
Their success suggests that scaling up the model size, training corpora and computational resources can yield models capable of solving a broad spectrum of tasks. 
However, beyond sheer scale, there is no evidence indicating that autoregressive modeling is the inevitable or optimal foundational paradigm for language modeling. Diffusion language models (DLMs)~\cite{khanna2025mercury,geminidiffusion,song2025seed} have recently emerged as a promising competitor, exhibiting parallel decoding (non-causal, high throughput), flexible orders, and potential for fewer inference steps. Built upon prior theoretical and empirical work~\cite{austin2021structured,lou2023discrete,nie2024scaling}, masked diffusion language models (MDLMs) have stood out from counterparts (i.e., DLMs) and established a dominant position. 
 
A representative train-from-scratch MDLM—LLaDA~\cite{nie2025large} demonstrates competitive performance compared to similarly sized AR models. Beyond the promising results, two problems regarding MDLMs remain underexplored: \textbf{(1) Decoding Strategy}, as shown in the (d) of Fig.~\ref{fig:motivation}, why does semi-AR (i.e., block-wise) decoding outperform full diffusion (without block constraint) decoding without specifically training for this strategy? 
The (a), (b) and (c) of Fig.~\ref{fig:motivation} may unveil the reason behind this behavior. In particular, (c.1) and (c.2) indicate two key characteristics: 1) The average tokens confidence is lower in the initial denoising steps but increases sharply as denoising progresses. 2) The average confidence of \EOS is significantly higher than that of \nonEOS. These characteristics explain the phenomena observed in (a) and (b). The block constraint avoids decoding the \EOS at the end of sequence in early steps, whereas full diffusion decoding lacks such constraint. Consequently, when employing full diffusion decoding, few semantically informative tokens are decoded in the initial low-confidence phase, potentially causing derailment—the model deviates from the correct denoising trajectory. 
We refer to this phenomenon the '\Trap' of full diffusion decoding. Furthermore, characteristic 1) suggests that uniform step-size decoding scheduler may be suboptimal. An intuitive approach is to decode cautiously in the early low-confidence phase and more aggressively in the later steps.
\textbf{(2) Online Reinforcement Learning}, several recent studies attempt to replicate the success of online reinforcement learning (RL) in AR models to MDLMs. For example, d1~\cite{zhao2025d1} directly applies GRPO~\cite{shao2024deepseekmath,guo2025deepseek} to LLaDA~\cite{nie2025large}, which overlooks the inconsistency between rollout trajectory and optimization trajectory.  
While LLaDOU~\cite{huang2025reinforcing} recognizes this issue and introduces an unmasking policy module to assist low confidence remasking, then combines it with RL to improve performance, this module requires training from scratch and often needs customized training for different model sizes and architectures, which is neither elegant nor scalable. Moreover, even when trajectory consistency considered, hardware memory confronts a significant bottleneck in online reinforcement learning. As maintaining consistency between rollout and optimization trajectories necessitates storing intermediate states throughout the denoising process, memory consumption drastically increases and becomes prohibitive as the number of denoising steps grows.

To this end, we propose approaches from the three perspectives, including decoding strategy, denoising step-size scheduler and reinforcement learning algorithm to alleviate the above problems. In summary, our contributions are three-fold:
\begin{itemize}
    \item \textbf{EOS Early Rejection (EOSER):} We introduce EOS early rejection (EOSER) for full diffusion decoding, which excessively suppress the confidence of \EOS in early steps and gradually restores it as denoising progresses, thereby alleviating the \Trap. Experimental results show that EOSER significantly improves the performance of full diffusion decoding.

    \item \textbf{Ascending Step-Size (ASS) Decoding Scheduler:} Based on the observation that average token confidence evolves from low to a sharply increasing during denoising, we propose an ascending step-size decoding scheduler instead of uniform step-size ones. This scheduler decodes fewer tokens cautiously in early steps and more tokens aggressively in later steps, substantially reducing the number of decoding steps (e.g., from $\frac{L}{2}$ to $\log_2 L$).

    \item \textbf{Consistency Trajectory Group Relative Policy Optimization (CJ-GRPO):} We propose Consistency Trajectory GRPO (CJ-GRPO) to align the denoising trajectories of rollout and optimization. It considers the non-causal nature of full diffusion decoding and effectively reduces the optimization errors caused by skip-step optimization. When combined with the EOSER and ASS scheduler, CJ-GRPO exhibits the potential to achieve considerable performance with fewer decoding steps.
\end{itemize}

\section{Related Work}
\label{sec:related work}

\subsection{Masked Diffusion Language Models}
\label{subsec:masked diffusion language models}
Recently, utilizing diffusion models to construct large language models (DLMs)~\cite{nie2025large,ye2025dream} has emerged as a promising trend. Many works~\cite{nie2025large,ye2025dream,austin2021structured,wu2023ar,sahoo2024simple,lou2023discrete,zheng2024masked,gong2024scaling,ou2024your,nie2024scaling,arriola2025block} are obsessed with the parallel decoding, flexible order, and potential fewer inference steps of DLMs, considering them a strong competitor to autoregressive models. 
D3PM~\cite{austin2021structured} initiates the discrete probability diffusion model, proposing four commonly used corruption approaches to damage sentences in the forward process: uniform, absorbing state (mask), discrete Gaussian, and token embedding distance. Among these, masked diffusion language models (MDLMs) have stood out and established a dominant position among the counterparts. Recent research has scale MDLMs to the billion-parameter regime, achieving performance comparable to similarly sized autoregressive LLMs (e.g., SMDM~\cite{gong2024scaling}, LLaDA~\cite{nie2025large}, Dream~\cite{ye2025dream}, DiffuLLaMA~\cite{nie2024scaling}), significantly scaling their capabilities. Other notable works include Diffusion-LM~\cite{li2022diffusion} and simple guidance~\cite{schiff2024simple} explore controllable text generation for DLMs. Block Diffusion~\cite{arriola2025block} introduces semi-autoregressive training-inference paradigm, combining intra-block diffusion, inter-block autoregression to enable variable-length generation.

\subsection{Online Reinforcement Learning for Diffusion Language Models}
\label{subsec:online reinforcement learning for diffusion language models}
Reinforcement learning (RL) has been extensively explored to align autoregressive (AR) LLMs on human preference and enhancing their reasoning capabilities~\cite{schulman2017proximal,rafailov2023direct,shao2024deepseekmath}.
More recently, MDLMs~\cite{ye2024diffusion,huang2025reinforcing,zhu2025llada,zhao2025d1} have shown promise in reasoning tasks (e.g., math, game). For instance, DoT~\cite{ye2024diffusion} demonstrates that DLMs are effective in producing chain-of-thought reasoning and addressing reasoning tasks. d1~\cite{zhao2025d1} directly adopts the representative online reinforcement learning algorithm (i.e., GRPO~\cite{shao2024deepseekmath,guo2025deepseek}) to adapt MDLMs, while LLaDOU~\cite{huang2025reinforcing} explores online reinforcement learning algorithms more suitable for DLMs/MDLMs.
So far, most policy gradient-based RL algorithms~\cite{shao2024deepseekmath,guo2025deepseek,zeng2025simplerl,yu2025dapo,hu2025open} have been tailored for autoregressive LLMs, which perform with causal mask and token-by-token generation. 
These algorithms are seamlessly compatible with sequential modeling and decision making. However, the non-causal attention nature of MDLMs prevents RL algorithms designed for AR models from being directly applied to MDLMs.
\section{Methodology}
\label{sec:methodology}

\subsection{Masked Diffusion Large Language Models}
\label{subsec:masked large language diffusion models}
Similar to typical diffusion models~\cite{ho2020denoising,song2020denoising}, MDLMs~\citep{austin2021structured,lou2023discrete,shi2024simplified} model a distribution $p_\theta(x_0)$ through a discrete forward and reverse process defined by $q(x_{t+\tau}|x_t)$ and $p_\theta(x_t|x_{t+\tau})$, respectively. The forward process gradually corrupts a sequence $x_0$ by replacing tokens with \MASK according to a time index $t \in [0, 1]$. At $t=0$, the sequence $x_0$ is clean and unmasked. The intermediate state $x_t$ is partially masked and each token is masked with the probability $t$ or unmasked with $1-t$ at time tick $t$. The reverse process recovers the original sequence $x_0$ from the fully masked state $x_1$.
During the pre-training, the model learns a mask predictor $p_\theta(\cdot|x_t)$ that takes $x_t$ as input and predicts all \MASK. The training objective minimizes the negative evidence lower bound, which upper-bounds the negative log-likelihood of the model distribution:
\begin{equation}
 - \mathbb{E}_
{p_{\text{data}}(x_0)} \left[ \log p_{\theta}(x_0) \right] \leq \mathcal{L}_\theta,
\label{eq:elbo}
\end{equation}
\begin{equation}
\mathcal{L}_\theta = -  \mathbb{E}_
{{t\sim \mathcal{U}(0,1]}, \; x_0 \sim p_{\text{data}}, \; x_t \sim q_{t|0}(x_t | x_0)} \left[ \frac{1}{t} \sum_{i}^{|x_t|} \mathbb{I}[x_t^i = \texttt{<MASK>}] \log p_{\theta}(x_0^i \mid x_t) \right],
\label{eq:pretraining_loss}
\end{equation}
where $|x_t|$ is the sequence length of $x$, $k$ means the $k$-th position, $\mathbb{I}$ denotes the indicator function that ensures the loss is computed only for masked tokens.
Instruction tuning follows a similar procedure: the prompt $p_0$ remains unchanged, while the response $r_0$ is partially masked. The model takes the prompt $p_0$ and masked response $r_t$ as input and is trained to predict the clean response $r_0$:
\begin{equation}
 \mathcal{L}'_\theta = -  \mathbb{E}_
{{t\sim \mathcal{U}(0,1]}, \; p_0 \sim p_{\text{data}}, \; r_t \sim q_{t|0}(r_t | p_0, r_0)} \left[ \frac{1}{t} \sum_{i=1}^{|r_t|} \mathbb{I}[r_t^i = \texttt{<MASK>}] \log p_{\theta}(r_0^i \mid p_0, r_t) \right],
\label{eq:sft_loss}
\end{equation}
where $|r_t|$ is the generation length of response.
In practice, the continuous time variable $t \in (0,1]$ is discretized into the step $\tilde{s}$ with interval $\Delta t = \frac{1}{S}$, such that $t=\tilde{s}\cdot\Delta t,\tilde{s} \in \{0,1,\cdots,S-1\}$. The denoising step $s$ is defined as $s=S- \tilde{s},s \in \{1,0,\cdots,S-1\}$, implying $x_s=x_{S-\tilde{s}}=x_{S-S\cdot t}=x_{1-t}$. For ease of description, we will use $s$ in the remainder of this paper.

\begin{figure}[t!]
  \centering
    \includegraphics[width=1.\textwidth]{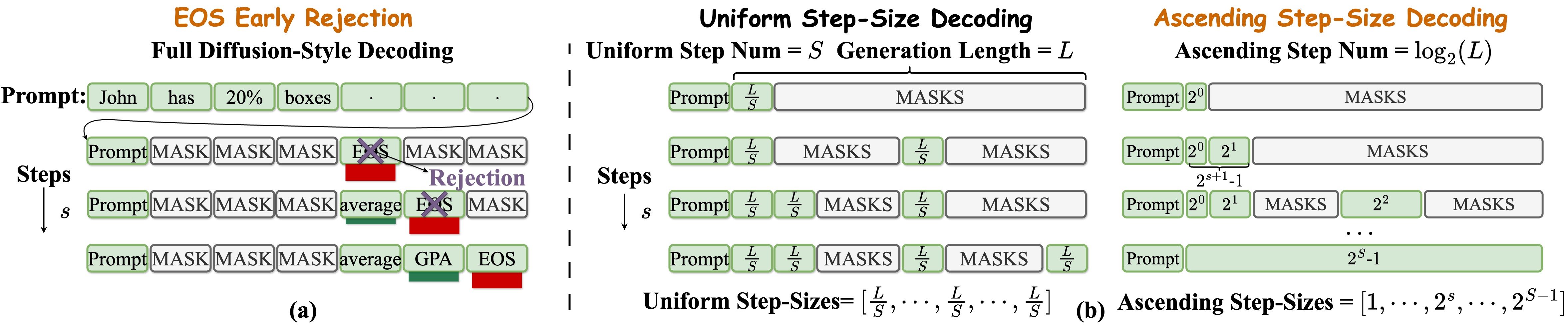}
    \vspace{-0.7cm}
    \caption{\textbf{(a)} EOS Early Rejection. \textbf{(b)} Ascending Step-Size Scheduler for fewer inference steps.}
    \label{fig:decoding}
\end{figure}

\subsection{EOS Early Rejection}
\label{subsec:eos early rejection}
As illustrated in Fig.~\ref{fig:motivation} (Sec.~\ref{sec:introduction}), the performance of full diffusion-style decoding significantly lags behind that of semi-autoregressive (block-wise) decoding. This issue arises because LLaDA~\cite{nie2025large} replaces \PAD with \EOS during pre-training, resulting the training sequences include a large number of \EOS, and enabling the model to learn such bias from data. Consequently, during the early steps of full diffusion decoding—when the average token confidence is low—the model exhibits disproportionately high confidence in \EOS compared to \nonEOS, leading it prematurely generate \EOS at the end of sequence in early steps. In contrast, commonly used block-wise decoding avoids the \Trap by isolating later blocks. However, it introduces several limitations: (1) The block boundaries restricts decoding flexibility. Tokens in later blocks that may have high confidence and should be decoded earlier are forcibly delayed, potentially constraining the optimal decoding trajectory. (2) Block length may be a sensitive hyperparameter that requires careful tuning. In comparison, full diffusion-style decoding is free from such constraints.

To alleviate the \Trap associated with full diffusion-style decoding, we introduce a attenuation coefficient $\gamma$ for \EOS's confidence that varies with the denoising steps $s$:
\begin{equation}
\label{eq:gamma}
 \gamma = \gamma_{\mathrm{min}} + (\gamma_{\mathrm{max}}-\gamma_{\mathrm{min}})\frac{s}{S-1},
\end{equation}
\begin{align}
\label{eq:eoser}
\hat{p}_\theta(x^i_s|x_{s-1}) = 
& \left\{
\begin{aligned} 
& \gamma \cdot p_\theta(x^i_s|x_{s-1}), \qquad \text{if  } \text{decode}(x^i_s)=\texttt{EOS}\\ 
& p_\theta(x^i_s|x_{s-1}),  \qquad \text{otherwise  } \\ 
\end{aligned},
\right.
\end{align}
here, $\gamma_{\mathrm{min}}$ and $\gamma_{\mathrm{max}}$ represent the lower and upper bound, respectively, \textit{s} is the current step, and \textit{S} is the total number of steps.
The attenuation coefficient $\gamma$ is used for early rejection mechanism. This mechanism suppresses the occurrence of \EOS in the early steps by multiplying the confidence corresponding to early \EOS by the attenuation coefficient $\gamma$, as shown in Eqn.~\ref{eq:eoser} and Fig.~\ref{fig:decoding} (a), while restoring its probability in the later steps to allow for sentence termination. Empirically, we set $\gamma_{max}=1.0$ and $\gamma_{min} \in [0.4,0.6]$ for uniform step-size decoding scheduler.

\begin{figure}[t!]
  \centering
    \includegraphics[width=1.\textwidth]{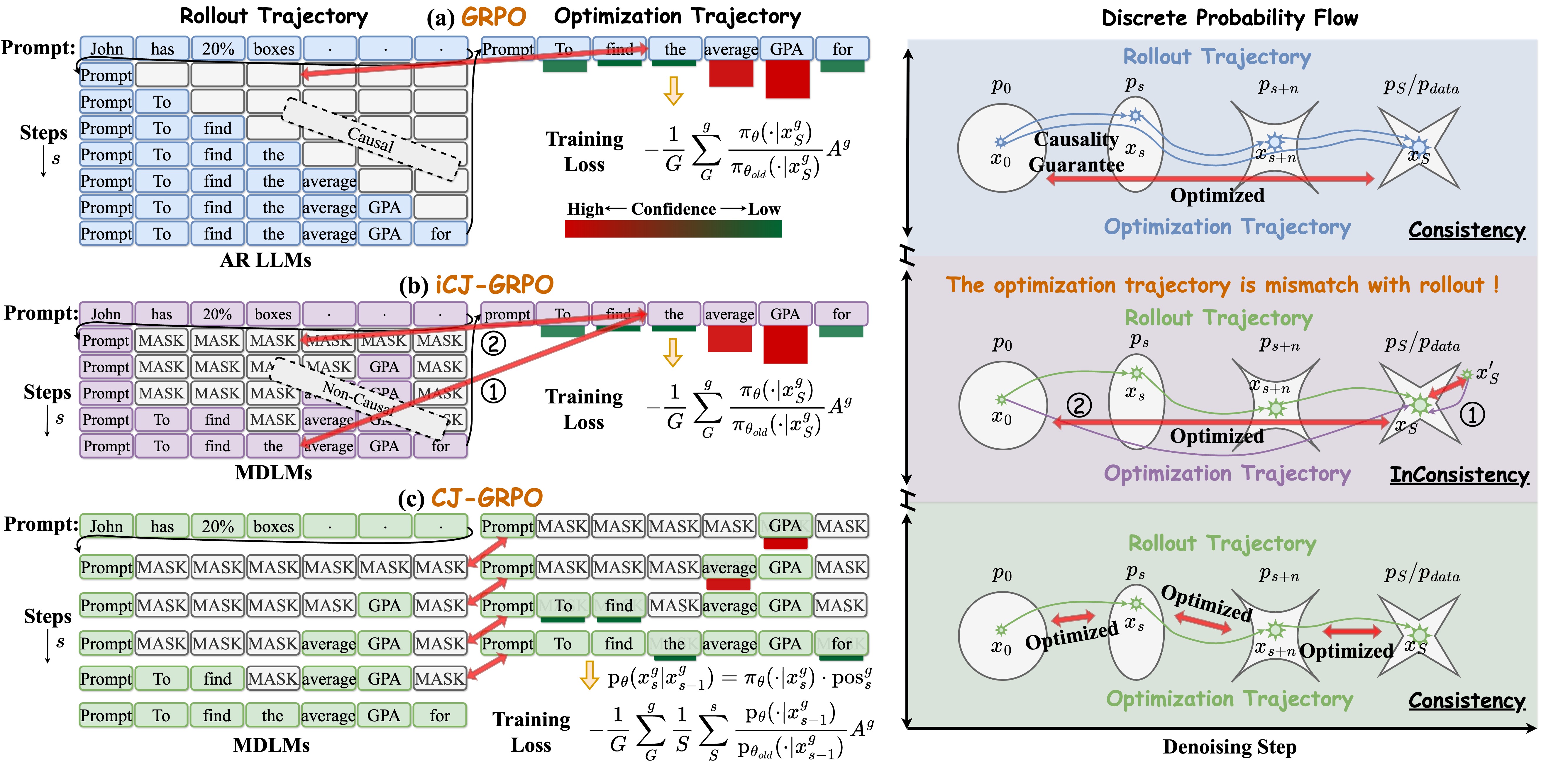}
    \vspace{-0.8cm}
    \caption{(a) The causality gurarantee the consistency between AR LLMs' rollout and optimization. (b) Two inconsistency trajectories RL training for MDLMs, we refer to $\textcircled{1}$ as one-step $x'_S \rightarrow x_S$ and $\textcircled{2}$ as one-step $x_0 \rightarrow x_S$. (c) Consistency Trajectory Group Relative Policy Optimization.}
    \label{fig:algorithm}
\end{figure}

\subsection{Ascending Step-Size Scheduler for Fewer Steps Inference}
\label{subsec:ass}
As shown in Fig.~\ref{fig:motivation} and discussed in Sec.~\ref{sec:introduction}, the average token confidence remains low in early denoising steps and increases sharply in later steps. This observation suggests that a cautious decoding strategy—denoising fewer tokens—is preferable initially, while a more aggressive approach—denoising more tokens—can be adopted later. To leverage this phenomenon, we propose an ascending step-size decoding scheduler that adapts to the confidence trend, significantly reducing the number of denoising steps required.
As illustrated in Fig.~\ref{fig:decoding} (b), we replace the fixed step-size of $\frac{L}{S}$ with a power-of-2 schedule to determine the number of tokens decoded at each step $s$. Specifically, at decoding step $s \in \{0,\cdots,s,\cdots,S-1\}$, the number of tokens decoded is $2^s$, and the cumulative number of tokens decoded up to step $s$ is $2^{s+1}-1$. The current and cumulative token counts are thus defined as: $\mathrm{cur\_tokens}=\{1,\cdots,2^s,\cdots,2^{S-1}\},\mathrm{cum\_tokens}=\{{1,\cdots,2^{s+1}-1,\cdots,2^S-1}\}.$
Since the sum of the first S terms of a geometric progression with a common ratio of 2 is $2^S-1$, but considering that the target generation length $L$ is typically set to a power of 2, we adjust the final step to decode one additional token. Thus the ascending step-size scheduler is defined as follows:
\begin{align}
\label{eq:ass_scheduler}
 \text{Ascending\, Step-Size\, Scheduler}(s) = 
& \left\{
\begin{aligned} 
& 2^s+1, \qquad \text{if  } s=S-1\\ 
& 2^s,  \qquad \text{otherwise  } \\ 
\end{aligned}.
\right.
\end{align}
To avoid the \Trap when using Ascending Step-Size (ASS) scheduler with full diffusion decoding, we integrate the EOSER. Given that the step-sizes are non-uniform, we adjust the $\gamma$ to account for the progressive increase in power-of-2. Specifically, the $\gamma_{\mathrm{ASS}}$ is defined as follows:
\begin{align}
\label{eq:ass_gamma}
 \gamma_{\mathrm{ASS}} = 
& \left\{
\begin{aligned} 
& \gamma_{\mathrm{min}} + (\gamma_{\mathrm{max}}-\gamma_{\mathrm{min}})\frac{2^s+1}{2^S}, \qquad \text{if  } s=S-1\\ 
& \gamma_{\mathrm{min}} + (\gamma_{\mathrm{max}}-\gamma_{\mathrm{min}})\frac{2^s}{2^S},  \qquad \text{otherwise  } \\ 
\end{aligned}.
\right.
\end{align}
Empirically, we set $\gamma_{max}=1.0$ and $\gamma_{min} = 0.01$. The ASS scheduler can also be extended to block-wise decoding. For example, multiple power-of-2 step groups can be combined into a single block. An example is as follows:
\begin{align}
\label{eq:block_ass_scheduler}
\text{Blocks} = \{ \underbrace{2^0,2^1,2^2,2^3,2^4,}_{\text{block 1, block length=$2^5-1$}}\cdots, \underbrace{2^s,2^{s+1},2^{s+2},2^{s+3},2^{s+4}}_{\text{block n, block length=$2^s\cdot(2^5-1)$}},\cdots\}.
\end{align}
The ASS decoding significantly reduces the inference steps (time) of MDLMs. Several studies~\cite{nie2025large,zhao2025d1} demonstrate that the empirically optimal uniform steps for MDLMs are often half the generation length $\frac{L}{2}$. The ASS decoding scheduler reduce this from $\frac{L}{2}$ to $\log_2 L$, thereby lowering the time complexity from \( \mathcal{O}(L) \) to \( \mathcal{O}(\log_2 L) \). This exhibits significantly potential for accelerating inference.

\subsection{Consistency Trajectory Group Relative Policy Optimization}
\label{subsec:cj-grpo}
GRPO~\cite{shao2024deepseekmath,guo2025deepseek} is a widely used and cost-effective online reinforcement learning algorithm for training autoregressive (AR) models. However, directly applying GRPO to masked diffusion language models (MDLMs) may lead to a fundamental mismatch. Unlike AR decoding, Semi-AR and full diffusion-style decoding in MDLM lacks the \textbf{causality guarantee} (i.e., causal attention). As depicted in (a) of Fig.~\ref{fig:algorithm}, in AR models, the predicted confidence of each token depends only on previously decoded tokens, and token confidences obtained during rollout are completely consistent with the confidences computed on final prompt-completion pairs.

In contrast, MDLMs employ bidirectional attention, leading to non-causal (diffusion-style) rollout trajectories. The predicted token confidences during rollout which are estimated in the presence of \MASK, differ from the confidences computed on final prompt-completion pair ($p_0,r_0$ in Eqn.\ref{eq:sft_loss}). This discrepancy arises because the model's predictions at intermediate steps are conditioned on partially masked sequences, unlike the autoregressive paradigm, as illustrated in Fig.~\ref{fig:algorithm} (a).
To address the issue posed by non-causal rollout trajectories in MDLMs, there are two compromise \textbf{i}n\textbf{c}onsistency tra\textbf{j}ectory GRPO algorithm (we refer to them as \textbf{iCJ-GRPO}) in Fig.\ref{fig:algorithm} (b). $\textcircled{1}$ involves calculating confidence from prompt-completion pairs with slightly perturbed prompts (as in d1~\cite{zhao2025d1}). This method can be seen as a one-step denoising optimization from a starting point $x'_S$ near the final state $x_S$, or $\textcircled{2}$ directly computes confidence from fully masked responses $r_1$($x_0$), and optimizing to reach the final answers $r_0$($x_S$), equivalent to performing the one-step denoising optimization from initial state $x_0$ to the final state $x_S$. Both of them may bring significant optimization errors of trajectory and may degrade performance.
To ensure consistency between rollout and optimization trajectories, we propose the \textbf{C}onsistency Tra\textbf{j}ectory \textbf{G}roup \textbf{R}elative \textbf{P}olicy \textbf{O}ptimization (\textbf{CJ-GRPO}), as illustrated in Fig.~\ref{fig:algorithm} (c). During the rollout phase, we record intermediate states at each denoising step $s$ in two queues: one for storing token confidences $\pi_{\theta}(\cdot|x^g_{s-1})$, and the other for storing the position $\mathrm{pos}^g_s$ of the token decoded at each step, which exhibit the following relationship:
\begin{equation}
\label{eq:confidence-pos}
\mathrm{p}_{\theta}(x^g_s|x^g_{s-1}) = \pi_{\theta}(\cdot|x^g_{s-1}) \cdot \mathrm{pos}^g_s,
\end{equation}
where $\mathrm{p}_{\theta}(x^g_s|x^g_{s-1})$ represents token confidences (probabilities) corresponding to the decoding positions $\mathrm{pos}^g_s$ at step $s-1$, and $g$ denotes the group ID in GRPO. These intermediate states are then utilized to guide the optimization trajectory.
The loss between adjacent steps $s-1 \rightarrow s$ is as follow:
\begin{equation}
\label{eq:trajectory-loss}
\mathcal{L}_{\theta,s-1} = -\frac{1}{G}\sum_{g=1}^G\frac{\mathrm{p}_{\theta}(x^g_s|x^g_{s-1})}{\mathrm{p}_{\theta_{\mathrm{old}}}(x^g_s|x^g_{s-1})}A^g + \beta \mathrm{KL}(\pi_\theta\|\pi_{\theta_{\mathrm{ref}}}).
\end{equation}
The sample-wise trajectory loss and the corresponding batched trajectory loss are:
\begin{equation}
\label{eq:batch-trajectory-loss}
\mathcal{L}^b_{\theta} = \frac{1}{S}\sum_{s=1}^S\mathcal{L}_{\theta,s-1}, \quad \mathcal{L}_{\theta}=\frac{1}{B} \sum_{b=1}^{B}\mathcal{L}^b_{\theta}.
\end{equation}
The detailed optimization pipeline is depicted in Algo.~\ref{alg:cj-grpo}.
However, a limitation of consistency trajectory GRPO algorithm is that storing intermediate states increases memory overhead proportionally with the number of denoising steps $S$. To address this issue, we naturally consider employing ASS scheduler during rollout. This not only reduces the time complexity from $\mathcal{O}(L)$ to $\mathcal{O}(\log_2{L})$, but also the space complexity from $\mathcal{O}(L)$ to $\mathcal{O}(\log_2{L})$. Furthermore, models trained in this manner exhibit improved fewer-step generation capability. The combination of CJ-GRPO and ASS scheduler brings multiple advantages simultaneously, embodying the \textit{'Kill three birds with one stone'}.

\begin{algorithm}[t!]
\small\small
\textbf{Input} Unmasking policy $\pi_\theta$, training dataset $\mathcal{D}$, reward verifier $\mathrm{R}(x,a)$, and queue of denoising trajectory $\mathrm{Q}(x_s,\mathrm{pos}_s)$ for recording intermediate states and unmasking positions.
\caption{CJ-GRPO training with ASS and EOSER}
\label{alg:cj-grpo}
\begin{algorithmic}[1]
\While{$\pi_\theta$ not converged or max epochs unreached}
    \State Sample a batch of question-answer pair $(q, a) \in \mathcal{D}$
    \For{$g$ \textbf{in} $1, ..., G$} \Comment{$G$ denotes the number of groups per question}
        \State Initialize $x^g_0$ with $q$ and mask tokens
        \For{$s$ \textbf{in} $1, ..., S$} \Comment{$S$ denotes the number of denoising steps}
        \State Decoding $x^g_s \sim \pi_{\theta}(\cdot|x^g_{s-1})$ with ASS \& EOSER
        \State Enqueue $\mathrm{Q}(x^g_{s-1}, \mathrm{pos}^g_s)$
        \EndFor
        \State $r^g=\mathrm{R}(x^g_S, a)$ \Comment{Compute the rewards}
    \EndFor
    \State $A^g = \frac{r^g-\mathrm{mean}(r^{1:G})}{\mathrm{std}(r^{1:G})}$ \Comment{Compute the advantages}   
    \For{$i$ \textbf{in} $1, ...,\mu$} \Comment{$\mu$ denotes the number of GRPO iterations} 
        \For{$s$ \textbf{in} $1, ..., S$} \Comment{Compute denoising trajectory loss}
            \State Dequeue $\mathrm{Q}(x^g_{s-1}, \mathrm{pos}^g_s)$
            \State $\mathrm{p}_{\theta}(x^g_s|x^g_{s-1}) = \pi_{\theta}(\cdot|x^g_{s-1}) \cdot \mathrm{pos}^g_s$, $\mathrm{p}_{\theta_{\mathrm{old}}}(x^g_s|x^g_{s-1}) = \pi_{\theta_{\mathrm{old}}}(\cdot|x^g_{s-1}) \cdot \mathrm{pos}^g_s$ 
            \State $\mathcal{L}_{\theta,s-1} = -\frac{1}{G}\sum_{g=1}^G\frac{\mathrm{p}_{\theta}(x^g_s|x^g_{s-1})}{\mathrm{p}_{\theta_{\mathrm{old}}}(x^g_s|x^g_{s-1})}A^g + \beta \mathrm{KL}(\pi_\theta\|\pi_{\theta_{\mathrm{ref}}})$   
            \Comment{see Eqn.~\ref{eq:trajectory-loss}}
        \EndFor
        \State 
        $\mathcal{L}^b_{\theta} = \frac{1}{S}\sum_{s=1}^S\mathcal{L}_{\theta,s-1}$, calculate the gradient $\mathcal{L}_{\theta}=\frac{1}{B} \sum_{b=1}^{B}\mathcal{L}^b_{\theta}$
        \Comment{see Eqn.~\ref{eq:batch-trajectory-loss}}
    \State Update $\theta$ of $\pi_\theta$ with $\nabla_\theta\mathcal{L}_{\theta}=\frac{1}{B} \sum_{b=1}^{B}\nabla_\theta\mathcal{L}^b_{\theta}$
    \EndFor
\EndWhile
\end{algorithmic}
\textbf{Output} Unmask policy $\pi_\theta$.
\end{algorithm}
\section{Experiments}
\label{sec:experiments}

Following previous work d1~\cite{zhao2025d1}, we evaluate LLaDA~\cite{nie2025large} after reinforcement learning on mathematical, planning, and coding tasks to assess how our proposed decoding strategies and reinforcement learning algorithm improve reasoning capabilities.

\subsection{Implementation Details}
\label{subsec:implementation details}
\paragraph{Datasets and Benchmarks}
We evaluate our method on following domains, (1) \textbf{Mathematics}: The mathematical problems which can be verified the correctness between model's response and ground truth answer with robust regular expressions. We utilize GSM8K~\cite{cobbe2021training}, a high-quality dataset of linguistically diverse grade school math word problems, and MATH500~\cite{lightman2023let}, a curated subset of 500 problems filtered from the MATH~\cite{hendrycks2021measuring}.
(2) \textbf{Planning}: Involving two categories of tasks: 4x4 Sudoku puzzles, which require constraint satisfaction and systematic elimination to fill a grid with numbers, and Countdown with 3 numbers, a combinatorial arithmetic game where models must reach a target number using basic arithmetic operations on a given set of numbers.

\paragraph{Model and Training}
We utilize LLaDA-8B-Instruct~\cite{nie2025large} as the base model. To ensure a fair comparison, we share the same configuration as diffu-GRPO~\cite{zhao2025d1}. For both diffu-GRPO and our CJ-GRPO, the group size for group-relative advantage estimation is 6, the KL-divergence coefficient $\beta$ is 0.04. For supervised fine-tuning (SFT), we train on the s1k~\cite{muennighoff2025s1} dataset for 20 epochs iteration, and the maximum sequence length of 4096. The importance sampling parameter $\epsilon$ is set to 0.5, and the temperature of Gumbel noise sampling~\cite{zheng2024masked} during rollout is set to 1.0. The number of GRPO iterations $\mu$ is set to 12 for math tasks and 8 for planning tasks.
We set the $\gamma_{max}=1.0$ for uniform and ASS decoding. For uniform decoding, $\gamma_{min} \in [0.4,0.6]$, while for ASS scheduler, $\gamma_{min}=0.01$.
All baseline methods (SFT, diffu-GRPO) and CJ-GRPO are trained with a batch size of 48 and 2 gradient accumulation steps. 

\paragraph{Evaluation and metric}
We compare our methods against SFT and diffu-GRPO~\cite{zhao2025d1} across all benchmarks. Evaluation is performed using 0-shot prompting and greedy decoding. The generation lengths are set to 128 and 256, with corresponding block lengths of 64 and 128 (i.e., half the generation length). The number of denoising steps is set to 7/8, 32/64, and 64/128 for the respective configurations.

\begin{table}[t] 
\vspace{-0.1cm}
\centering 
\caption{\textbf{Performance Comparison on Countdown, GSM8K, MATH500 and Sudoku:} We report results under different decoding strategies and training algorithms, with generation lengths $L$ of 128 and 256 across denoising steps $S$=$\frac{L}{4}$ and $\frac{L}{2}$ with uniform step-size. The notation '$+$' indicates a combination of decoding strategies and algorithms, '*' denotes training-free, and \sethlcolor{gray!25}\hl{gray cell} represents our methods. \textbf{Bold} font indicates the optimal while \underline{underline} is the suboptimal performance. }
\vspace{-0.2cm}
\setlength{\arrayrulewidth}{0.9pt}
\setlength{\cmidrulewidth}{0.9pt} 
\scalebox{0.59}{
\begin{tabular}{l cc|cc| cc|cc| cc|cc| cc|cc} 
\toprule[1.15pt]
& \multicolumn{4}{c|}{\textbf{Countdown}} & \multicolumn{4}{c|}{\textbf{GSM8K}} & \multicolumn{4}{c|}{\textbf{MATH500}} & \multicolumn{4}{c}{\textbf{Sudoku}} \\ 
\cmidrule(){2-17}
\textbf{Generation Length $L$} & \multicolumn{2}{c|}{\textbf{128}} & \multicolumn{2}{c|}{\textbf{256}} & \multicolumn{2}{c|}{\textbf{128}} & \multicolumn{2}{c|}{\textbf{256}} & \multicolumn{2}{c|}{\textbf{128}} & \multicolumn{2}{c|}{\textbf{256}} & \multicolumn{2}{c|}{\textbf{128}} & \multicolumn{2}{c}{\textbf{256}}\\
\cmidrule(lr){2-5} \cmidrule(lr){6-9} \cmidrule(lr){10-13} \cmidrule(lr){14-17}
\textbf{Methods / Steps $S$} & \textbf{32} & \textbf{64} & \textbf{64} & \textbf{128} & \textbf{32} & \textbf{64} & \textbf{64} & \textbf{128} & \textbf{32} & \textbf{64} & \textbf{64} & \textbf{128} & \textbf{32} & \textbf{64} & \textbf{64} & \textbf{128} \\
\midrule[1.15pt]
Semi-AR* & 26.56 & 30.08 & 30.47 & 19.92 & \textbf{65.58} & \textbf{70.20} & \textbf{63.31} & \textbf{76.65} & \textbf{26.20} & \textbf{26.60} & \textbf{30.00} & \textbf{31.20} & 6.88 & 11.62 & \textbf{9.42} & \textbf{2.64} \\ 
Full Diffusion* & 35.94 & 5.08 & 0.00 & 0.00 & 56.94 & 51.86 & 34.87 & 22.97 & 22.20 & 20.40 & 19.60 & 15.60 & 5.76 & 8.98 & 0.00 & 0.83 \\ 
\rowcolor{gray!25}
EOSER* & \textbf{44.14} & \textbf{46.88} & \textbf{41.80} & \textbf{39.45} & 60.73 & 63.15 & 58.45 & 61.33 & 22.20 & 24.60 & 22.40 & 24.20 & \textbf{13.18} & \textbf{13.92} & 2.69 & 1.42 \\ 
\midrule[1.15pt]
SFT & 23.83 & 30.47 & 18.36 & 24.22 & 61.49 & 68.39 & 61.87 & 77.48 & 23.80 & 27.40 & 23.00 & 31.40 & 6.54 & 8.84 & 11.82 & 4.44 \\ 
diffu-GRPO & 35.16 & 28.52 & 24.22 & 28.52 & \underline{64.59} & \underline{71.72} & 65.58 & \underline{79.91} & \underline{28.20} & \underline{31.20} & \underline{30.40} & \underline{33.80} & 24.41 & 22.66 & 19.04 & 18.85 \\ 
\rowcolor{gray!25}
CJ-GRPO + Semi-AR & \underline{58.20} & \underline{61.33} & \underline{63.52} & \underline{67.81} & \textbf{77.48} & \textbf{78.39} & \textbf{80.41} & \textbf{84.29} & \textbf{28.80} & \textbf{33.20} & \textbf{33.40} & \textbf{38.00} & \underline{27.83} & \underline{54.10} & \underline{47.90} & \underline{57.80} \\ 
\rowcolor{gray!25}
CJ-GRPO + EOSER & \textbf{66.02} & \textbf{70.80} & \textbf{69.34} & \textbf{75.59} & 62.85 & 67.10 & \underline{71.72} & 77.75 & 23.60 & 23.20 & 26.60 & 28.40 & \textbf{85.25} & \textbf{85.69} & \textbf{85.96} & \textbf{85.37} \\ 
\bottomrule[1.15pt]
\end{tabular}
}
\label{tab:main results}
\vspace{-0.3cm}
\end{table}

\begin{figure}[t!]
\vspace{-15pt}
    \centering
    \begin{minipage}[t]{0.67\textwidth}
     \vspace{0pt}
        \centering
        \begin{table}[H]
            \centering
            \caption{\textbf{Performance of fewer decoding steps on Countdown, GSM8K, MATH500 and Sudoku:} We report results under various decoding strategies and training algorithms when the decoding steps are set to $\log_2 L$. For ascending step-size scheduler, the steps are set to $\log_2 L$, whereas for uniform step-size ones, we fix the steps at 8 to ensure divisibility with both 128 and 256.}
            \vspace{-0.2cm}
            \setlength{\arrayrulewidth}{0.9pt}
            \setlength{\cmidrulewidth}{0.9pt} 
            \scalebox{0.55}{
                \begin{tabular}{l cc| cc| cc| cc} 
                    \toprule[1.15pt]
                    & \multicolumn{2}{c|}{\textbf{Countdown}} & \multicolumn{2}{c|}{\textbf{GSM8K}} & \multicolumn{2}{c|}{\textbf{MATH500}} & \multicolumn{2}{c}{\textbf{Sudoku}} \\
                    \cmidrule(lr){2-9}
                    \textbf{Generation Length $L$} & \textbf{128} & \textbf{256} & \textbf{128} & \textbf{256} & \textbf{128} & \textbf{256} & \textbf{128} & \textbf{256} \\
                    \cmidrule(lr){2-3}\cmidrule(lr){4-5} \cmidrule(lr){6-7} \cmidrule(lr){8-9}
                    \textbf{Methods / Steps $S$} & \textbf{7/8} & \textbf{8} & \textbf{7/8} & \textbf{8} & \textbf{7/8} & \textbf{8} & \textbf{7/8} & \textbf{8} \\ 
                    \midrule[1.15pt]
                    Semi-AR* + Uniform ($S=8$) & 18.75 & 4.69 & 42.15 & 26.61 & 17.20 & 13.20 & \textbf{6.05} & 0.59 \\ 
                    Full Diffusion*  + Uniform ($S=8$) & 11.33 & 0.00 & 44.12 & 12.74 & 17.00 & 9.20 & 0.00 & 0.44 \\ 
                    \rowcolor{gray!25}
                    EOSER* + Uniform ($S=8$) & 5.86 & \underline{22.27} & 43.44 & \underline{29.42} & 16.20 & 14.20 & 0.05 & 0.20 \\ 
                    \rowcolor{gray!25}
                    Semi-AR* + ASS ($S=\log_2 L$) & 2.34 & 0.00 & 35.10 & 9.40 & 12.80 & 14.00 & \underline{4.44} & \underline{1.03} \\ 
                    \rowcolor{gray!25}
                    Full Diffusion* + ASS ($S=\log_2 L$) & \underline{19.14} & 0.00 & \underline{46.55} & 17.97 & \underline{19.20} & 12.20 & 3.22 & 0.00 \\ 
                    \rowcolor{gray!25}
                    EOSER* + ASS ($S=\log_2 L$) & \textbf{29.30} & \textbf{51.17} & \textbf{48.14} & \textbf{50.80} & \textbf{20.00} & \textbf{19.60} & 3.17 & \textbf{18.26} \\ 
                    \midrule[1.15pt]
                    SFT + Semi-AR + Uniform ($S=8$) & 28.91 & 22.66 & 25.25 & 12.28 & 12.00 & 6.40 & 8.64 & 0.73 \\ 
                    diffu-GRPO + Semi-AR + Uniform ($S=8$) & \underline{30.08} & \underline{26.56} & 32.83 & \underline{20.32} & 12.00 & 18.40 & 21.09 & 7.03 \\
                    \rowcolor{gray!25}
                    CJ-GRPO + Semi-AR + ASS ($S=\log_2 L$) & 3.12 & 1.95 & \underline{38.44} & 5.53 & \underline{16.40} & \underline{19.20} & \underline{24.95} & \underline{19.04} \\
                    \rowcolor{gray!25}
                    CJ-GRPO + EOSER + ASS ($S=\log_2 L$) & \textbf{53.91} & \textbf{59.38} & \textbf{55.57} & \textbf{53.22} & \textbf{21.20} & \textbf{22.20} & \textbf{43.31} & \textbf{50.32} \\ 
                    \bottomrule[1.15pt]
                \end{tabular}
            }
            \label{tab:fewer step results}
        \end{table}
    \end{minipage}
    \hfill
    \begin{minipage}[t]{0.3\textwidth}
    \vspace{12pt}
        \centering
        \includegraphics[width=0.99\textwidth]{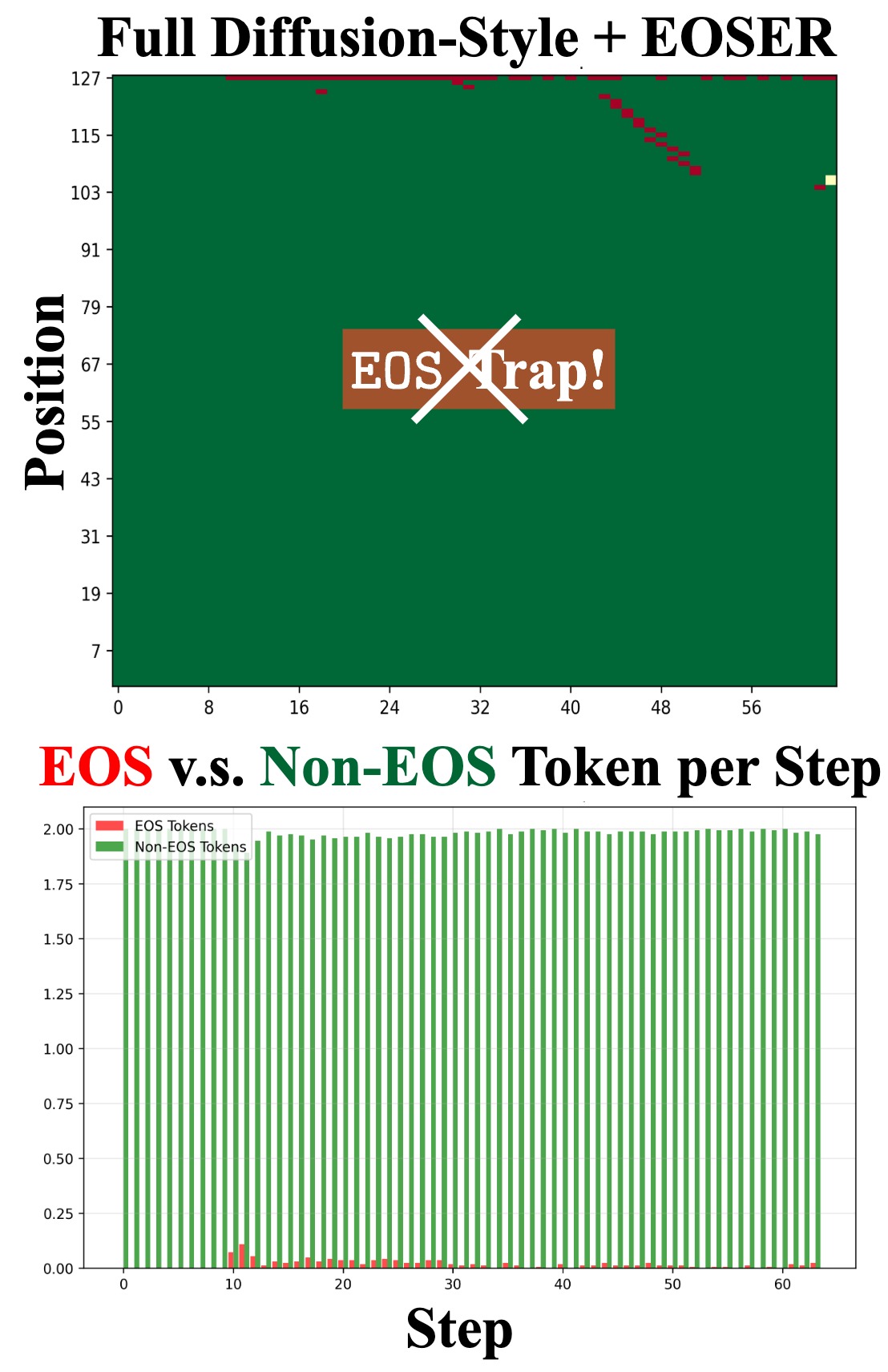}
            \vspace{-0.7cm}
        \caption{The Step-Position Heatmap of EOSER.} 
        \label{fig:EOSER}
    \end{minipage}
\vspace{-2pt}
\end{figure}


\subsection{Main Results}
\label{subsec:main results}
We report the performance of baseline LLaDA-8B-Instruct~\cite{nie2025large} equipped with different decoding strategies —training-free comparison—in the upper part of Tab.~\ref{tab:main results} and Tab.~\ref{tab:fewer step results}. The lower sections of the same tables show the results on various post-training recipes across four tasks, evaluated in a zero-shot setting where each model is trained specifically for each task.

\textbf{Training-free Setting Comparison Among Decoding Strategies.} 
We compare the performance of three decoding strategies under the training-free setting: semi-autoregressive (AR), full diffusion and EOSER. For semi-AR decoding, the block length is set to half the generation length, $\frac{L}{2}$, 
based on empirical studies that show this configuration yields better performance. In contrast, for full diffusion and EOSER, the block length is equal to the generation length, $L$. 
The results are illustrated in the upper part of Tab.~\ref{tab:main results} and the Fig.~\ref{fig:EOSER}. 
Our findings are as follows: (1) Full diffusion decoding performs significantly worse than the other two strategies across all tasks. In the planning tasks, EOSER outperforms semi-AR, particularly on the Countdown task. 
However, in mathematics questions, EOSER is inferior to semi-AR. We provide a detailed analysis of this in Sec.~\ref{subsec:analysis}. (2) EOSER effectively mitigates the \Trap (i.e., \EOS are not decoded per step, leading to a large number of \EOS being ejected in the sequence) and significantly improve the performance of full diffusion. Fig.~\ref{fig:EOSER} indicates that the arrival of \EOS is delayed both spatially (in position) and temporally (in step).

\textbf{CJ-GRPO Combined with Semi-AR and EOSER Compared to SFT and diffu-GRPO.} 
We evaluate the performance of CJ-GRPO combined with semi-AR and EOSER decoding separately, as shown in Tab.~\ref{tab:main results}. We can observe that CJ-GRPO+Semi-AR outperforms both SFT and diffu-GRPO across all tasks and 16 settings. Specifically, for planning tasks, Countdown and Sudoku, CJ-GRPO+Semi-AR achieves performance levels that are twice as high as those of diffu-GRPO in several configurations. Moreover, CJ-GRPO+EOSER significantly surpasses SFT, diffu-GRPO, and even CJ-GRPO+Semi-AR on planning tasks. The improvements are particularly notable on Sudoku, where CJ-GRPO+EOSER achieves performance that is approximately twice that of CJ-GRPO+Semi-AR and four times that of diffu-GRPO.
Interestingly, on the mathematics tasks, CJ-GRPO+EOSER underperforms compared to diffu-GRPO and is significantly lower than CJ-GRPO+Semi-AR. This finding is consistent with the results observed in the training-free settings. We provide a detailed analysis of these findings in Sec.~\ref{subsec:analysis}.

\textbf{Fewer Decoding Steps Potential of Ascending Step-Size (ASS) Scheduler and CJ-GRPO Combined with ASS.}  
We evaluate the performance of SFT, diffu-GRPO, and CJ-GRPO using fewer decoding steps, specifically $\log_2 L$, as depicted in Tab.~\ref{tab:fewer step results}. 
Under the training-free settings, the combination of EOSER+ASS demonstrates optimal performance across almost all benchmarks on 7 settings, outperforming various decoding methods and step-size schedulers. This result indicates that EOSER+ASS holds significant potential for reducing the number of decoding steps. 
Next, we train the model using CJ-GRPO combined with EOSER and Semi-AR (note that Semi-AR is also compatible with ASS scheduler as mentioned in Sec.~\ref{subsec:ass}). We found that even with only 7/8 ($\log_2 L$) steps compared to original 32/64/128 ($\frac{L}{2}$), EOSER+ASS achieves considerable performance, significantly surpassing other methods using either uniform step-size or the ASS scheduler in all settings. Regarding planning tasks, CJ-GRPO+EOSER+ASS even surpasses the original diffu-GRPO which using $\frac{L}{2}$ denoising steps. Please refer to the \textit{Appendix} for more discussion.

\subsection{Ablation Study}
\label{subsec:ablation study}
\begin{wraptable}{r}{7.5cm}
\vspace{-0.6cm}
\centering
\caption{Ablation on denoising trajectory optimizations.}
\vspace{-0.3cm}
\hspace{-0.3cm}
\setlength{\arrayrulewidth}{1.15pt}
\resizebox{0.54\textwidth}{!}{
\begin{tabular}{lcccc}
\toprule[1.15pt]
& \textbf{Countdown} & \textbf{GSM8K} & \textbf{MATH500} & \textbf{Sudoku} \\
\midrule[1.15pt]
$\textcircled{1}$ One-step $x'_S$ to $x_S$ & 35.16 & 64.59 & 28.20 & 24.41 \\
\midrule
$\textcircled{2}$ One-step $x_0$ to $x_S$ & 45.31 & 68.76 & 25.00 & 25.68 \\
\midrule
\rowcolor{gray!25}
\makecell{\rowcolor{gray!25} Consistency Trajectory \\ $[x_0,\cdots,x_s,\cdots,x_S]$} & \textbf{58.20} & \textbf{77.48} & \textbf{28.80} & \textbf{27.83} \\
\bottomrule[1.15pt]
\end{tabular}
}
\vspace{-0.3cm}
\label{tab:ablation}
\end{wraptable}

\textbf{Ablation on Consistency v.s. Inconsistency in Trajectory Optimization.} 
We conduct an ablation study to evaluate the two trajectory optimization methods introduced in Sec.~\ref{subsec:cj-grpo} to clarify the effect of consistent versus inconsistent trajectory optimization. Experiments are conducted with a generation length 128 ($L$), a block length 64 ($\frac{L}{2}$), and 32 denoising steps ($\frac{L}{4}$). 
We refer to $\textcircled{1}$ as one-step from $x'_S$ to $x_S$ (prompt masking as used in d1~\cite{zhao2025d1}), and $\textcircled{2}$ as one-step from $x_0$ to $x_S$, and ours as consistency trajectory optimization over $[x_0,\cdots,x_s,\cdots,x_S]$. Results show that optimizing over each intermediate denoising step leads to significantly better performance.
Skipping intermediate steps and implementing $\textcircled{2}$ introduces substantial optimization errors of the actual rollout trajectory which leads to optimization biases to some extent. Moreover, implementing $\textcircled{2}$ is equivalent to optimize from a point $x'_S$ near the final solution $x_S$, represents a compromised alternative and yields the worst performance.

\subsection{Analysis}
\label{subsec:analysis}
\textbf{Analyzing of Characteristics of MDLMs on Reasoning Tasks.} As noted in Sec.~\ref{subsec:main results}, training-free and training-based (CJ-GRPO) EOSER decoding significantly outperforms Semi-AR decoding on planning tasks, yet slightly underperforms on mathematical tasks. We attribute this discrepancy to the inherent pattern differences between two types of reasoning tasks, namely planning and math. Specifically, Countdown exhibits a clear parallel reasoning pattern, while math have sequential pattern. 
For example, Countdown requires inferring missing operators sandwiched between given operands and target values, while Sudoku demands reasoning over both row and column constraints (2D grid), or thinking forward and backward from the unfilled space (1D sequence) simultaneously. 
In contrast, math problems in GSM8k and MATH500 typically follow a sequential reasoning process, where each step builds directly on the previous one. This pattern makes semi-AR decoding more suitable for mathematics. 
These observations suggest the value of future work that hybridizes diffusion-style (parallel) and autoregressive (sequential) reasoning to better adapt MDLMs to tasks with diverse reasoning patterns.

\section{Conclusion}
\label{sec:conclusion}

This paper introduces a reinforcement learning algorithm for MDLMS combined with a full diffusion-style decoding strategy and variable step-size scheduler. The goal is to tame MDLMs via reinforcement learning with fewer decoding steps. We propose three key components: (1) EOS Early Rejection (EOSER), which alleviates the \Trap in full diffusion-style decoding by suppressing early \EOS generation, (2) Ascending Step-Size (ASS) Decoding Scheduler, which reduces decoding steps while maintaining performance by adapting to the change trend of token confidence—low in early steps and sharply increasing later, and (3) Consistency Trajectory Group Relative Policy Optimization (CJ-GRPO), which aligns rollout and optimization trajectories to ensure the consistency. Experimental results on reasoning tasks demonstrate that CJ-GRPO is particularly effective and efficient for taming MDLMs. When combined with EOSER decoding and ASS scheduler, it achieves comparable performance to $\frac{L}{2}$ steps using only $\log_2 L$ steps.



\section{Ethics Statement}
\label{sec:ethics statement}
Our research focuses on reinforcement learning and decoding strategies for diffusion large language models. We do not involve human subjects in our study. The datasets used in our research are publicly available and do not contain any sensitive or private information. We have ensured that our research practices comply with all relevant legal regulations. We have no conflicts of interest or sponsorship to disclose. Our research aims to improve the performance and efficiency of diffusion large language models, and we believe that our methods and findings are ethical and do not raise any concerns regarding discrimination, bias, fairness, privacy, security, or research integrity.

\section{Reproducibility Statement}
\label{sec:reproducibility statement}
Ensuring the reproducibility of our work is a priority. In the experimental section of our paper, we provide detailed descriptions of the parameter settings and experimental configurations for the reinforcement learning and decoding strategies applied to masked diffusion large language models—LLaDA-8B-Instruct. To further support reproducibility, we have open-sourced our codebase to the community. This codebase includes all necessary scripts and instructions for reproducing our results. Additionally, any datasets used in our experiments are publicly accessible. We believe that these efforts will enable other researchers to replicate our findings and build upon our work.

\bibliography{iclr2026_conference}
\bibliographystyle{iclr2026_conference}


\newpage
\section{Appendix}
\appendix
\textbf{Roadmap}

\label{sec: roadmap}
In the Appendix, we present the supplementray experiments and discussions in App.~\ref{sec:supplementary experiments and discussions}, the limitations in App.~\ref{sec:limitations}, and the use of Large Language Models (LLMs) statement in App.~\ref{sec:the use of LLMs}.


\section{Supplementary Experiments and Discussions}
\label{sec:supplementary experiments and discussions}

\begin{table}[h!] 
\centering 
\caption{\textbf{Ablation study on the denoising steps: Countdown.} We report the performance changes as denoising steps varies from $1$ to $256,2^s,$ across decoding strategies. '*' denotes training-free.
}
\setlength{\cmidrulewidth}{0.9pt}
\scalebox{0.85}{
\begin{tabular}{l ccccccccc} 
\toprule[1.15pt]
\textbf{Gen Len $L=256$} & \multicolumn{9}{c}{\textbf{Countdown}} \\ 
\cmidrule(lr){2-10}
\textbf{Denoising Steps} & \textbf{1} & \textbf{2} & \textbf{4} & \textbf{8} & \textbf{16} & \textbf{32} & \textbf{64} & \textbf{128} & \textbf{256} \\ 
\midrule[1.15pt]
Semi-AR* (Block Len $N=64$)& - & - & 0.00 & 4.69 & 23.83 & 16.02 & 15.62 & \textbf{27.34} & 21.09 \\ 
Full-Diffusion* & 0.00 & 0.00 & 0.00 & 0.00 & \textbf{0.78} & 0.00 & 0.00 & 0.00 & 0.00 \\  
EOSER* & 0.00 & 0.00 & 8.98 & 22.27 & 21.48 & 23.83 & 38.67 & \textbf{41.80} & 39.45 \\ 
\bottomrule[1.15pt]
\end{tabular}
}
\label{tab:ablation1}
\end{table}

\begin{table}[h!] 
\centering 
\caption{\textbf{Ablation study on the denoising steps: GSM8K.} We report the performance changes as denoising steps varies from $1$ to $256,2^s,$ across decoding strategies. '*' denotes training-free.
}
\setlength{\cmidrulewidth}{0.9pt}
\scalebox{0.85}{
\begin{tabular}{l ccccccccc} 
\toprule[1.15pt]
\textbf{Gen Len $L=256$} & \multicolumn{9}{c}{\textbf{GSM8K}} \\ 
\cmidrule(lr){2-10}
\textbf{Denoising Steps} & \textbf{1} & \textbf{2} & \textbf{4} & \textbf{8} & \textbf{16} & \textbf{32} & \textbf{64} & \textbf{128} & \textbf{256} \\ 
\midrule[1.15pt]
Semi-AR* (Block Len $N=64$)& - & - & 16.76 & 26.61 & 30.78 & 48.14 & 71.27 & 77.71 & \textbf{79.98} \\ 
Full-Diffusion* & 3.11 & 3.49 & 3.94 & 12.74 & 36.85 & \textbf{42.46} & 34.34 & 34.87 & 22.97 \\  
EOSER* & 3.11 & 16.22 & 23.28 & 29.42 & 38.59 & 47.76 & 52.08 & 58.45 & \textbf{61.33} \\ 
\bottomrule[1.15pt]
\end{tabular}
}
\label{tab:ablation2}
\end{table}

\begin{table}[h!] 
\centering 
\caption{\textbf{Ablation study on the denoising steps: MATH500.} We report the performance changes as denoising steps varies from $1$ to $256,2^s,$ across decoding strategies. '*' denotes training-free.
}
\setlength{\cmidrulewidth}{0.9pt}
\scalebox{0.85}{
\begin{tabular}{l ccccccccc} 
\toprule[1.15pt]
\textbf{Gen Len $L=256$} & \multicolumn{9}{c}{\textbf{MATH500}} \\ 
\cmidrule(lr){2-10}
\textbf{Denoising Steps} & \textbf{1} & \textbf{2} & \textbf{4} & \textbf{8} & \textbf{16} & \textbf{32} & \textbf{64} & \textbf{128} & \textbf{256} \\ 
\midrule 
Semi-AR* (Block Len $N=64$)& - & - & 6.40 & 13.20 & 18.60 & 22.40 & 27.60 & 33.20 & \textbf{37.80} \\ 
Full-Diffusion* & 6.00 & 6.00 & 6.40 & 9.20 & 16.40 & 19.80 & \textbf{21.20} & 17.80 & 17.20 \\  
EOSER* & 6.00 & 9.60 & 12.20 & 14.20 & 18.20 & 20.20 & 22.40 & \textbf{24.20} & 22.67 \\ 
\bottomrule[1.15pt]
\end{tabular}
}
\label{tab:ablation3}
\end{table}

\begin{table}[h!] 
\centering 
\caption{\textbf{Ablation study on the denoising steps: Sudoku.} We report the performance changes as denoising steps varies from $1$ to $256,2^s,$ across decoding strategies. '*' denotes training-free.
}
\setlength{\cmidrulewidth}{0.9pt}
\scalebox{0.85}{
\begin{tabular}{l ccccccccc} 
\toprule[1.15pt]
\textbf{Gen Len $L=256$} & \multicolumn{9}{c}{\textbf{Sudoku}} \\ 
\cmidrule(lr){2-10}
\textbf{Denoising Steps} & \textbf{1} & \textbf{2} & \textbf{4} & \textbf{8} & \textbf{16} & \textbf{32} & \textbf{64} & \textbf{128} & \textbf{256} \\ 
\midrule[1.15pt]
Semi-AR* (Block Len $N=64$)& - & - & 0.05 & 0.59 & 2.44 & 5.22 & \textbf{10.84} & 5.42 & 5.66 \\ 
Full-Diffusion* & 0.00 & 0.00 & 0.00 & 0.44 & 8.20 & \textbf{8.54} & 3.96 & 0.68 & 0.44 \\  
EOSER* & 0.00 & 0.15 & 0.15 & 0.20 & 0.10 & 2.10 & 2.64 & \textbf{2.69} & 1.42 \\ 
\bottomrule[1.15pt] 
\end{tabular}
}
\label{tab:ablation4}
\end{table}

\textbf{Ablation on Denoising Steps.}  
As depicted in Tab.~\ref{tab:ablation1}, ~\ref{tab:ablation2}, \ref{tab:ablation3}, \ref{tab:ablation4}. We conduct ablation studies on the number steps $S$ using different decoding strategies across four benchmarks. We observe a common trend: for Semi-AR and EOSER, the best performance is often achieved when the number of denoising steps is half the generation length $\frac{L}{2}$. Regarding the full diffusion-style decoding strategy, the optimal number of steps is usually smaller than $\frac{L}{2}$, additionally, while for EOSER (also full diffusion-style), although its optimal number of step is $\frac{L}{2}$, the performance difference across denoising steps such as 64, 128, and 256 is relatively small, however, for Semi-AR decoding strategy, block length may be a sensitive hyperparameter, which may result in a large performance variance obtained. This indicates that full diffusion-style decoding has greater potential for fast generation (fewer decoding steps). Specifically, when the generation length is $L$ and the number of denoising step $\frac{L}{2}$, two tokens are generated at each decoding step. This does not significantly increase the token throughput compared to the next-token prediction of AR models (one-by-one). Moreover, the presence of fixed-length \MASK means that the generation speed of MDLMs with more decoding steps (e.g. uniform step-size scheduler) may not surpass that of AR models with the same software and hardware infrastructure.
Our findings indicate that exploring few-step generation for full diffusion-style decoding in MDLMs is meaningful. Semi-AR, while effective, is not the ultimate solution. Finally, this exploration may truly reveal the speed advantages of MDLMs compared to AR LLMs. Currently, the performance improvements are not significant, and the speed advantages are not fully realized.

\section{Limitations}
\label{sec:limitations}
In Sec.\ref{subsec:cj-grpo}, we discussed a limitation faced by CJ-GRPO: as the number of steps increases, so does the amount of memory required, because storing intermediate states increases memory overhead proportionally with the number of denoising steps $S$. This constraint greatly affects the training speed. To mitigate this issue, we already have proposed the Ascending Step Size (ASS) scheduler. In planning tasks, the combination of CJ-GRPO with EOSER and ASS, which generates fewer steps ($\log_2 L$), has proven to be even more effective than the previous uniform step-size method that relies on more denoising steps ($\frac{L}{2}$). This not only reduces the time complexity from $\mathcal{O}(L)$ to $\mathcal{O}(\log_2{L})$, but also the space complexity from $\mathcal{O}(L)$ to $\mathcal{O}(\log_2{L})$. 
However, when it comes to mathematical tasks, this approach still falls short compared to diffu GRPO+Semi-AR and CJ-GRPO+Semi-AR. Given these observations, we believe that exploring better fewer-step full diffusion-style decoding strategies in the future holds great promise. Such methods could potentially resolve the memory issues associated with CJ-GRPO and enhance its mathematical reasoning capabilities. This area of research is well worth exploring.

\section{The Use of Large Language Models (LLMs) Statement}
\label{sec:the use of LLMs}
In the process of drafting this paper, we utilized large language models (LLMs) as a general-purpose assist tool. Specifically, the LLM was employed to enhance the clarity and coherence of the content, refine the logical structure of the arguments, and improve the grammatical accuracy of the text. The LLM provided suggestions and revisions that helped in polishing the overall quality of the manuscript. However, it is important to note that the LLM did not play a significant role in the research ideation or the core writing of the scientific content. The ideas, methodologies, and conclusions presented in this paper are the original work of the authors. We take full responsibility for the content of this paper, including any text that has been refined using the LLM.

\newpage
\section{Showcase on Sequential and Parallel Reasoning Tasks }
\label{sec:showcase}
Our analysis of the model's reasoning characteristics (patterns) in mathematical and planning tasks is primarily intuitive and draws loose analogies to human cognitive processes, without rigorous formalization. Therefore, the following cases and interpretations should be considered preliminary and illustrative rather than definitive.

\begin{tcolorbox}[title=\textbf{Sequential Reasoning Pattern: GSM8K}]

\textbf{Question:} Emmett does 12 jumping jacks, 8 pushups, and 20 situps. What percentage of his exercises were pushups?
\\
\\
\\
\textbf{Response:}

$<$reasoning$>$

To determine the percentage of Emmett's exercises that were pushups, \textcolor{purple}{we first must follow these steps}:

\qquad \textcolor{purple}{1. Calculate the total number of exercises Emmett did.}

\qquad \textcolor{purple}{2. Identify the number of pushups Emmett did.}

\qquad \textcolor{purple}{3. Determine the ratio of pushups to the total number of exercises.}

\qquad \textcolor{purple}{4. Convert the ratio to a percentage.}
\\

\textcolor{purple}{1. Given:}

   \qquad - Jumping jacks: 12
   
   \qquad - Pushups: 8
   
   \qquad - Situps: 20

   The total number of exercises is:
   \[ 
   12 + 8 + 20 = 40 
   \]

\textcolor{purple}{2. The number of pushups is 8.}

\textcolor{purple}{3. The ratio of pushups to the total number of exercises is:}
   \[ 
   \frac{8}{40} 
   \]

\textcolor{purple}{4. Simplify the ratio:}
   \[
   \frac{8}{40} = 0.2 
   \]

\textcolor{purple}{5. Convert the ratio to a percentage:}
   \[ 
   0.2 \times 100 = 20 
   \]
   
$<$/reasoning$>$

$<$answer$>$

\textcolor{green!70!black}{\[\boxed{20\%}\]}

$<$/answer$>$
\end{tcolorbox}

Mathematical reasoning tasks typically exhibit a sequential structure, where each step logically follows the previous one. For an example of GSM8K, which involving the calculation of a pushups percentage. The denominator is obtained by summing the given numerators, after which the final percentage is derived through division. The \textcolor{purple}{purple font} represents the sequential reasoning pattern among different sub-steps.

\newpage
\begin{tcolorbox}[title=\textbf{Sequential Reasoning Pattern: MATH500}]

\textbf{Question:} 
You are a math expert. You will be given a question to solve. Solve it step by step. Wrap the final answer in a \textbackslash boxed\{\}.
Container A holds 4 red balls and 6 green balls; containers B and C each hold 6 red balls and 4 green balls. A container is selected at random and then a ball is randomly selected from that container. What is the probability that the ball selected is green? Express your answer as a common fraction.
\\
\\
\\
\textbf{Response:}

$<$reasoning$>$

\textcolor{purple}{1. Calculate the sum of probabilities of selecting each container:}

   \qquad - Probability of selecting each container: 
   \[
   \frac{1}{3}
   \]

\textcolor{purple}{2. Calculate the probability of selecting a green ball from each container:}

   \qquad - Probability of selecting a green ball from container A: 
   \[
   \frac{6}{10} = \frac{3}{5}
   \]
   
   \qquad - Probability of selecting a green ball from container B: 
   \[\frac{4}{10} = \frac{2}{5}\]
   
   \qquad - Probability of selecting a green ball from container C:    
   \[
   \frac{4}{10} = \frac{2}{5}
   \]

\textcolor{purple}{3. Calculate the overall probability of selecting a green ball:}

   \qquad - Probability of green ball: 
   \[
   \frac{1}{3} \times \frac{3}{5} + \frac{1}{3} \times \frac{2}{5} + \frac{1}{3} \times \frac{2}{5} = \frac{1}{3} \times \frac{7}{5} = \frac{7}{15}
   \]
   
$<$/reasoning$>$

$<$answer$>$

\textcolor{green!70!black}{\[\boxed{\frac{7}{15}}\]}

$<$/answer$>$
\end{tcolorbox}

For instance, in a MATH500 problem requiring the calculation of a ratio, one must first determine the numerator and denominator. Given individual numerators, the denominator is obtained by summing them. Calculating the proportions in each container separately. Then, since there are three containers—A, B, and C—the final result is derived by combining their respective probabilities. The \textcolor{purple}{purple font} represents the sequential reasoning pattern among different sub-steps.

\newpage
\begin{tcolorbox}[title=\textbf{Parallel Reasoning Pattern: Countdown}]

\textbf{Question:} Using only the numbers [49, 55, 53], create an arithmetic expression that evaluates to exactly 51. You must use all numbers from the list, and each number must be used exactly once. You may use the operations +, -, *, and / as needed. After reasoning, provide only your final expression inside $<$answer$>$$<$/answer$>$ tags without including an equals sign or the target number. For example, if the numbers are [2, 3, 4] and the target is 5, a valid answer is: $<$answer$>$\textbackslash n2*4-3\textbackslash n$<$/answer$>$.
\\
\\
\\
\textbf{Response:}

$<$reasoning$>$

To create an arithmetic expression that evaluates to 51 using the numbers 49, 55, and 53, we need to consider the operations +, -, *, and /. Let's start by trying different combinations of these numbers:

\[
1. \textcolor{blue}{49} \textbf{+} \textcolor{blue}{55} \textbf{-} \textcolor{blue}{53} = \textcolor{green!90!black}{51}
\]

\[
2. \textcolor{blue}{49} \textbf{+} \textcolor{blue}{53} \textbf{-} \textcolor{blue}{55}  = \textcolor{red!90!black}{47}
\]

\[
3. \textcolor{blue}{55} \textbf{+} \textcolor{blue}{53} \textbf{-} \textcolor{blue}{49} =  \textcolor{red!90!black}{59}
\]

After trying different combinations, we find that the expression works:

\[
1. 49 + 55 - 53 = 51
\]

So, the expression 49 + 55 - 53 evaluates to 51.
   
$<$/reasoning$>$

$<$answer$>$

\textcolor{green!70!black}{\[\boxed{49 + 55 - 53}\]}

$<$/answer$>$
\end{tcolorbox}

The Countdown game is naturally amenable to parallel reasoning, as inferring an operator between two operands does not require strictly sequential steps. For example, the model need not determine operators between earlier operand pairs before handling subsequent ones. In the case illustrated (with operands highlighted in \textcolor{blue}{blue} and operators in \textbf{bold}), the model does not attempt pairwise combinations first, instead, it directly applies the $+$ and $-$ operation to three numbers—adding the first two and subtracting the third—and successfully concludes the correct answer.

\newpage
\begin{tcolorbox}[title=\textbf{Parallel Reasoning Pattern: Sudoku}]

\textbf{Question:} Please solve the following 4x4 Sudoku puzzle. The puzzle is provided as a 16-character string reading left-to-right, top-to-bottom, where '0' represents empty cells.

Rules:

\qquad - Fill empty cells with digits 1-4

\qquad - Each row must contain digits 1-4 exactly once

\qquad - Each column must contain digits 1-4 exactly once

\qquad - Each 2x2 box must contain digits 1-4 exactly once

Important: 

Your solution must be a COMPLETE 16-character string with only the digits 1-4, representing your final solved grid.

Respond in this exact format:

$<$reasoning$>$

Your step-by-step solving process

$<$/reasoning$>$

$<$answer$>$

[16-character solution string with no spaces or separators]

$<$/answer$>$ 

Solve the following Sudoku puzzle: \textbf{0000100420013142}.
\\
\\
\\
\textbf{Response:}

$<$reasoning$>$

To solve the given 4x4 Sudoku puzzle, we need to fill the empty cells with digits 1-4 such that each row, column, and 2x2 box contains all the digits exactly once. Here is the step-by-step process:

\qquad \textcolor{blue}{1. Identify Empty Cells: Locate all empty cells represented by '0' in the puzzle.}

\qquad \textcolor{blue}{2. Check Rows: Ensure each row contains the digits 1-4 exactly once.}

\qquad \textcolor{blue}{3. Check Columns: Ensure each column contains the digits 1-4 exactly once.}

\qquad \textcolor{blue}{4. Check 2x2 Boxes: Ensure each 2x2 box contains the digits 1-4 exactly once.}

\qquad \textcolor{blue}{5. Fill Empty Cells: Fill the empty cells with digits 1-4 while maintaining the above constraints.}

\qquad \textcolor{blue}{6. Verify the Solution: After filling the puzzle, verify that each row, column, and 2x2 box contains digits 1-4 exactly once.}

By following these steps, we can solve the given 4x4 Sudoku puzzle.
   
$<$/reasoning$>$

$<$answer$>$

\textcolor{green!70!black}{\[\boxed{\textbf{4213132424313142}}\]}

$<$/answer$>$
\end{tcolorbox}

Sudoku requires reasoning over both row and column constraints (in 2D grids), or simultaneous forward and backward thinking along sequences (in 1D formulations). This task is not well-suited to purely sequential reasoning but instead benefits from parallel—or bidirectional—reasoning. Accordingly, the model does not proceed by filling numbers step-by-step; rather, it first enumerates the relevant constraints (shown in blue), then deduces the answer directly. In this context, parallel reasoning can be viewed as an implicit process.

\end{document}